\documentclass{article}

\PassOptionsToPackage{numbers, compress}{natbib}

\usepackage[preprint]{neurips_2019}

\usepackage[utf8]{inputenc} 
\usepackage[T1]{fontenc}    
\usepackage{hyperref}       
\usepackage{url}            
\usepackage{booktabs}       
\usepackage{amsfonts}       
\usepackage{nicefrac}       
\usepackage{microtype}      
\usepackage{times}
\usepackage{bm}
\newtheorem{remark}{Remark}
\usepackage{epsfig}
\usepackage{graphicx}
\usepackage{amsmath}
\usepackage{amssymb}
\usepackage{subfigure}
\usepackage{pifont}
\usepackage[ruled]{algorithm2e}
\usepackage{algpseudocode}

\usepackage{latexsym}

\makeatletter
 
\def\ie{\emph{i.e}.} 
 
\def\etc{\emph{etc}.}

\makeatother

\title{Self-supervised Contrastive Attributed Graph Clustering}

\author{%
  Wei Xia\\
  Xidian University\\
  \And
  Quanxue Gao\thanks{The corresponding author: Quanxue Gao (E-mail: qxgao@xidian.edu.cn)} \\
  Xidian University\\
  \And
  Ming Yang\\
  Westfield State University\\
  \And
  Xinbo Gao\\
  Chongqing University of Posts and Telecommunications\\
}

\begin{document}

\maketitle

\begin{abstract}
  Attributed graph clustering, which learns node representation from node attribute and topological graph for clustering, is a fundamental but challenging task for graph analysis. Recently, methods based on graph contrastive learning (GCL) have obtained impressive clustering performance on this task. Yet, we observe that existing GCL-based methods 1) fail to benefit from imprecise clustering labels; 2) require a post-processing operation to get clustering labels; 3) cannot solve out-of-sample (OOS) problem. To address these issues, we propose a novel attributed graph clustering network, namely \textbf{\underline{S}}elf-supervised \textbf{\underline{C}}ontrastive \textbf{\underline{A}}ttributed \textbf{\underline{G}}raph \textbf{\underline{C}}lustering (SCAGC). In SCAGC, by leveraging inaccurate clustering labels, a self-supervised contrastive loss, which aims to maximize the similarities of intra-cluster nodes while minimizing the similarities of inter-cluster nodes, are designed for node representation learning. Meanwhile, a clustering module is built to directly output clustering labels by contrasting the representation of different clusters. Thus, for the OOS nodes, SCAGC can directly calculate their clustering labels. Extensive experimental results on four benchmark datasets have shown that SCAGC consistently outperforms 11 competitive clustering methods.
\end{abstract}

\section{Introduction}
\noindent In the era of Internet, network-structured data has penetrated into every corner of life. Representative examples include shopping networks~\cite{DataN}, social networks~\cite{PiaoZXCL21}, recommendation systems~\cite{HuangXXDXLBXLY21}, citation networks~\cite{WanPY021}, \etc. Real-world scenarios such as these can be modeled as attributed graphs, \ie, topological graphs structure with node attributes (or features). Due to non-Euclidean topological graph structure and complex node attribute, most existing machine learning approaches cannot be directly applied to analyze such data. To this end, graph neural networks (GNNs)~\cite{KipfW17} arises at the historic moment and have made great development in recent years. GNN aims to learn low-dimensional node representation for downstream tasks via simultaneously encoding the topological graph and node attribute. In this article, we will study the attributed graph clustering problem, which is one of the most challenging tasks in the fields of AI.

\begin{figure*}[!t]
	\centering
	\includegraphics[width=1.0\linewidth]{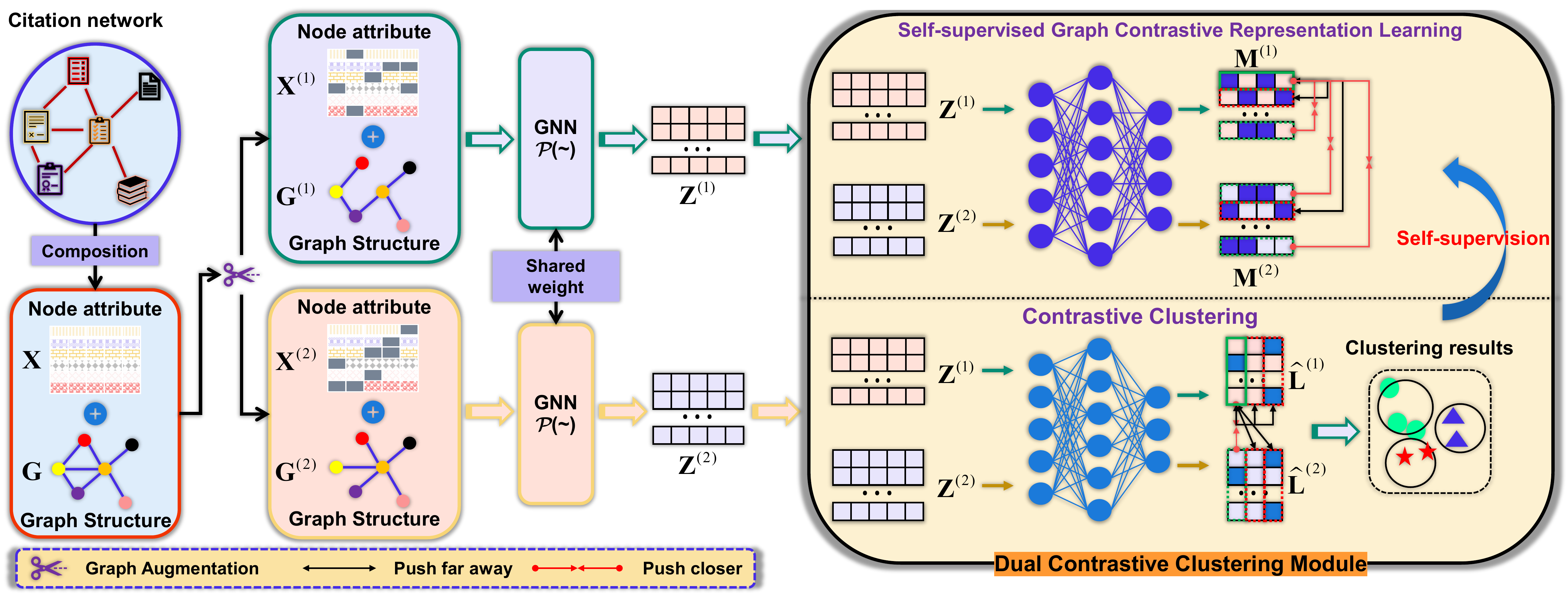}
	\caption{The framework of the proposed \textbf{\underline{S}}elf-supervised \textbf{\underline{C}}ontrastive \textbf{\underline{A}}ttributed \textbf{\underline{G}}raph  \textbf{\underline{C}}lustering (SCAGC).}
	\label{fig1}
\vspace{-4mm}
\end{figure*}

Attributed graph clustering, \ie, node clustering, aims to divide massive nodes into several disjoint clusters without intense manual guidance. To date, numerous attributed graph clustering methods have been proposed~\cite{WangPLZJ17,0003LLW19,ParkLCLC19,ChengWTXG20,FanWSLLW20,TMMnew,LinK21}, among which, most of them are based on graph auto-encoder (GAE) and variational GAE (VGAE)~\cite{KipfW16a}. For example, to learn a robust node representation, the variants of GAE and VGAE are proposed by~\cite{PanHLJYZ18,PanHFLJZ20}, namely adversarially regularized graph auto-encoder (ARGA) and adversarially regularized variational graph auto-encoder (ARVGA).
To build a clustering-directed network, inspired by deep embedding clustering (DEC)~\cite{XieGF16},~\cite{WangPHLJZ19} minimized the mismatch between clustering distribution and target distribution to improve the quality of node representation, and proposed deep attentional embedded graph clustering (DAEGC) approach. Similarly,~\cite{Bo0SZL020} presented structural deep clustering network (SDCN) to embed the topological structure into deep clustering. SDCN used the traditional auto-encoder to get new node feature via encoding node attribute, and then used GNN to simultaneously encode topological structure and new node feature to learn final node representation for clustering.~\cite{TuZL0CZC21} proposed deep fusion clustering network (DFCN), which used a dynamic cross-modality fusion mechanism for obtaining consensus node representation, thereby generating more robust target distribution for network optimizing. Although aforementioned methods have made encouraging progress, how to mine the highly heterogeneous information embedded in the attribute graph remains to be explored.

Recently, due to its powerful unsupervised representation learning ability, contrastive learning (CL) has made vast inroads into computer vision community~\cite{ChenK0H20,He0WXG20}. Motivated by this, several recent studies~\cite{VelickovicFHLBH19,SunHV020,abs-2106-09244,QiuCDZYDWT20,YouCSCWS20,XYLWW21,JinZL00P21,ZhaoYWYD21} show promising results on unsupervised graph representation learning (GRL) using approaches related to CL, we call this kind of methods graph contrast representation learning methods (GCRL for short in this paper). For example,~\cite{VelickovicFHLBH19} proposed deep graph information maximization (DGI) to learn node representation by contrasting the local node-level representation and the global graph-level representation. Similarly,~\cite{SunHV020} proposed to learn graph-level representation by maximizing the mutual information between the graph-level representation and representations of substructures. Based on the contrastive loss in SimCLR~\cite{ChenK0H20},~\cite{YouCSCWS20} proposed a new graph contrastive learning network with kinds of graph augmentation approaches (GraphCL) for facilitating node representation learning. More recently,~\cite{XYLWW21} first used adaptive graph augmentation schemes to construct different graph views, then extracted node representation via maximizing the agreement of node representation between graph views.

Though driven by various motivations and achieved commendable results, many existing GCRL methods still have the following challenging issues:
\begin{enumerate}
  \item They are task-agnostic, thus, will need a post-processing to get clustering labels, resulting in suboptimal node representation for down-stream node clustering task.
  \item They fail to benefit from imprecise clustering labels, thus suffering from inferior performances.
  \item They cannot handle out-of-sample (OOS) nodes, which limits their application in practical engineering.
\end{enumerate}

As shown in Figure~\ref{fig1}, we propose the self-supervised contrastive attributed graph clustering (SCAGC), a new attributed graph clustering approach that targets at addressing aforementioned limitations. In SCAGC, we first leverage graph augmentation methods to generate abundant attributed graph views, then, each augmented attributed graph has two compact representations: a clustering assignment probability produced by the clustering module and a low-dimension node representation produced by graph representation learning module. The two representations interact with each other and jointly evolve in an end-to-end framework. Specifically, the clustering module is trained via contrastive clustering loss to maximize the agreement between representations of the same cluster. The graph representation learning module is trained using the proposed self-supervised contrastive loss on pseudo labels, \ie, clustering labels, where nodes within a same cluster are trained to have similar representations. We perform experiments on four attributed graph datasets and compare with 11 state-of-the-art GRL and GCRL methods. The proposed SCAGC substantially outperforms all baselines across all benchmarks. The main contribution of the proposed SCAGC is two-fold:
\begin{enumerate}
  \item To the best of our knowledge, SCAGC could be the first contrastive attributed graph clustering work without post-processing. SCAGC can directly predict the clustering assignment result of given unlabeled attributed graph. For OOS nodes, SCAGC can also directly calculate the clustering labels without retraining the entire attributed graph, which accelerates the implementation of SCAGC in practical engineering.
  \item By benefiting form the clustering labels, we propose a new self-supervised CL loss, which facilitates the graph representation learning. Extensive experimental results witness its effectiveness for attributed clustering.
\end{enumerate}

\section{Methodology}
In this section, we first formalize the node clustering task on attributed graphs. Then, the overall framework of the proposed SCAGC will be introduced. Finally, we detail each component of the proposed network.

\subsection{Problem Formalization}
Given an arbitrary attributed graph $\mathcal{G}=(\textbf{U},\textbf{E},\textbf{X})$, where $\textbf{U}=\{\emph{\textrm{u}}_\textrm{1}\textrm{, \emph{u}}_\textrm{2}\textrm{, } \cdots\textrm{, \emph{u}}_\textrm{\emph{N}}\}$ is the vertex set, $\textbf{E}$ is the edge set, $\textbf{X}\in \mathbb{R}^{\emph{\textrm{N}}\times \emph{\textrm{d}}}$ is the node attribute matrix, $\textrm{\emph{N}}$ is the number of nodes, and $\textrm{\emph{d}}$ is the dimension of node attribute matrix. $\textbf{G}\in \mathbb{R}^{\emph{\textrm{N}}\times \emph{\textrm{N}}}$ is the adjacency matrix of $\mathcal{G}$, and $\textrm{G}_{\emph{ij}}=\textrm{1}$ iff $(\emph{\textrm{u}}_\emph{\textrm{i}}\textrm{, }\emph{\textrm{u}}_\textrm{\emph{j}})\in \textbf{E}$, \emph{i.e.}, there is an edge from node $\emph{\textrm{u}}_\emph{\textrm{i}}$ to $\emph{\textrm{u}}_\textrm{\emph{j}}$.

In this article, we study one of the most representative downstream tasks of GNNs, \emph{i.e.}, node clustering. The target of node clustering is to divide the given $\textrm{\emph{N}}$ unlabeled nodes into $\emph{\textrm{K}}$ disjoint clusters $\{\textbf{C}_\textrm{1, }\cdots\textrm{, }\textbf{C}_\textrm{\emph{k}}\textrm{, }\cdots\textrm{, }\textbf{C}_\emph{\textrm{K}}\}$, such that the node in the same cluster $\textbf{C}_\emph{\textrm{k}}$ has high similarity to each other~\cite{CuiZY020,TMMnew}.

\subsection{Overall Network Architecture}
As shown in Figure~\ref{fig1}, the network architecture of the proposed SCAGC consists of the following joint optimization components: shared graph convolutional encoder, contrastive clustering module and self-supervised graph contrastive representation learning module.
\begin{itemize}
  \item \textbf{Shared Graph Convolutional Encoder}: It aims to simultaneously map the augmented node attribute and topological graph structure to a new low-dimensional space for downstream node clustering task.
  \item \textbf{Self-Supervised GCRL Module}: To learn more discriminative graph representation and utilize the useful information embedded in inaccurate clustering labels, this module is designed to maximize the similarities of intra-cluster nodes, \emph{i.e.}, positive pairs, while minimizing the similarities of inter-cluster nodes, \emph{i.e.}, negative pairs.
  \item \textbf{Contrastive Clustering Module}: To directly get clustering labels, this module builds a clustering network by contrasting the representation of different clusters.
\end{itemize}

\subsection{Shared Graph Convolutional Encoder}
Graph contrastive representation has attracted much attention, due to its ability to utilize graph augmentation schemes to generate positive and negative node pairs for representation learning~\cite{YouCSCWS20,XYLWW21}. Specifically, given an arbitrary attributed graph $\mathcal{G}$ with node attribute $\textbf{X}$ and topological graph $\textbf{G}$, two stochastic graph augmentation schemes $\emph{\textbf{A}}^{\textrm{(1)}}\thicksim \bm{\mathcal{A}}$ and $\emph{\textbf{A}}^{\textrm{(2)}}\thicksim \bm{\mathcal{A}}$ are leveraged to construct two correlated attributed graph views \{$\textbf{X}^{\textrm{(1)}}\textrm{, } \textbf{G}^{\textrm{(1)}}$\} and \{$\textbf{X}^{\textrm{(2)}}\textrm{, }\textbf{G}^{\textrm{(2)}}$\}, where $\textbf{X}^{\textrm{(\emph{v})}}=\emph{\textbf{A}}^{\textrm{(\emph{v})}}\textrm{(\textbf{X})}$, and $\textbf{G}^{\textrm{(\emph{v})}}=\emph{\textbf{A}}^{\textrm{(\emph{v})}}\textrm{(\textbf{G})}$, $\emph{\textrm{v}}=\{\textrm{1, 2}\}$ is the $\emph{\textrm{v}}$-th graph augmentation, $\bm{\mathcal{A}}$ denotes the set of all kinds of graph augmentation methods, including attribute masking, edge perturbation. To be specific, attribute masking randomly adds noise to node attributes, and edge perturbation randomly adds or drops edges in topological graph. The underlying prior of these two graph augmentation schemes is to keep the intrinsic topological structure and node attribute of attributed graph unchanged. Based on this prior, the learned node representation will be robust to perturbation on insignificant attributes and edges. In this article, we implement the graph augmentations following the setting in GCA~\cite{XYLWW21}.

After obtaining two augmented attributed graph views \{$\textbf{X}^{\textrm{(1)}}\textrm{, } \textbf{G}^{\textrm{(1)}}$\} and \{$\textbf{X}^{\textrm{(2)}}\textrm{, }\textbf{G}^{\textrm{(2)}}$\}, we utilize a shared two-layer graph convolutional network $\bm{\mathcal{P}}(\thicksim)$ to simultaneously encode node attributes and topological graphs of augmented attributed graph views. Thus, we have
\begin{equation}\label{1}
\begin{aligned}
\overline{\textbf{Z}}^{\textrm{(\emph{v})}}&=\bm{\mathcal{P}}(\textbf{X}^{\textrm{(\emph{v})}}\textrm{,}\textbf{G}^{\textrm{(\emph{v})}}|\bm{\Omega}^\textrm{1})=\sigma( {\widetilde{\textbf{D}}_{\textrm{(\emph{v})}}^{-\frac{\textrm{1}}{\textrm{2}}}}\widetilde{\textbf{G}}^{\textrm{(\emph{v})}}{\widetilde{\textbf{D}}_{\textrm{(\emph{v})}}^{-\frac{\textrm{1}}{\textrm{2}}}}\textbf{X}^{\textrm{(\emph{v})}}\bm{\Omega}^\textrm{1} )\textrm{,}
\end{aligned}
\end{equation}

\begin{equation}\label{2}
\begin{aligned}
\textbf{Z}^{\textrm{(\emph{v})}} &= \bm{\mathcal{P}}(\overline{\textbf{Z}}^{\textrm{(\emph{v})}}\textrm{, }\textbf{G}^{\textrm{(\emph{v})}}|\bm{\Omega}^\textrm{2})\textrm{,}
\end{aligned}
\end{equation}
where $\overline{\textbf{Z}}^{\textrm{(\emph{v})}}$ is the 1-st layer's output of shared GNN; $\textbf{Z}^{\textrm{(\emph{v})}}\in \mathbb{R}^{\emph{\textrm{N}}\times \emph{\textrm{d}}_\textrm{1}}$ is the node representation under the $\emph{\textrm{v}}$-th graph augmentation; $\bm{\Omega}=\{\bm{\Omega}^\textrm{1}\textrm{, }\bm{\Omega}^\textrm{2}\}$ denotes the trainable parameter of graph convolutional encoder; $\widetilde{\textbf{G}}^{\textrm{(\emph{v})}} = \textbf{G}^{\textrm{(\emph{v})}} + \textbf{I}$; $\widetilde{\textbf{D}}^{\textrm{(\emph{v})}}\textrm{(\emph{i\emph{, }i})}=\sum_\textrm{\emph{j}}{\widetilde{\textbf{G}}_{\emph{\textrm{ij}}}^{\textrm{(\emph{v})}}}$; $\textbf{I}$ is an identity matrix; $\sigma\textrm{(}\cdot\textrm{)}=\textrm{max}\textrm{(}\textrm{\textrm{0, }}\textrm{)}$ represents the nonlinear ReLU activation function.

So far, we have obtained the node representations $\textbf{Z}^{\textrm{(1)}}$ and $\textbf{Z}^{\textrm{(2)}}$ of two augmented attributed graph views.
\begin{figure}[!t]
	\centering
	\includegraphics[width=0.5\linewidth]{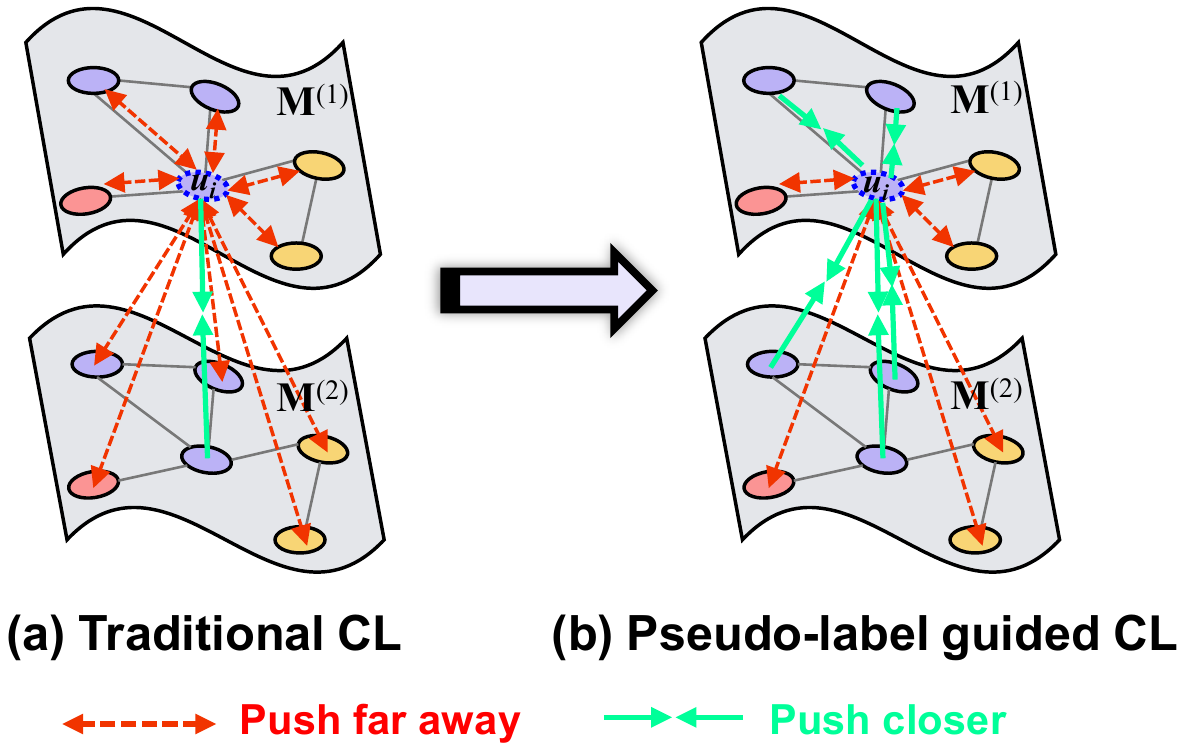}
	\caption{The illustration of self-supervised CL. Taking the node $\emph{\textrm{u}}_\textrm{\emph{i}}$ as an example, the nodes in the same cluster have the same color. In (a), we find that traditional CL mistakenly regards the remaining four positive nodes (purple nodes) in $\textbf{M}^{\textrm{(1)}}$ and $\textbf{M}^{\textrm{(1)}}$ as negative nodes of $\emph{\textrm{u}}_\textrm{\emph{i}}$.}
	\label{figill}
\vspace{-4mm}
\end{figure}

\subsection{Self-Supervised GCRL Module}
In the field of GRL, contrastive learning based GRL has been an effective paradigm for maximizing the similarities of positive pairs while minimizing the similarities of negative pairs to learn discriminative graph representation. For a given attributed graph with $\emph{\textrm{N}}$ nodes, there are $\textrm{2\emph{N}}$ augmented nodes. Traditional CL regard the representations of a node under two different augmentation as a positive pair, and leave other $\textrm{2\emph{N}-2}$
pairs to be negative (see \textbf{Figure~\ref{figill} (a)}). While having promising performance, this assumption runs counter to the criterion of clustering. In node clustering, \textbf{\emph{we hope that the nodes in the same cluster $\emph{\textbf{C}}_{\textrm{k}}$ have high similarity to each other while the nodes in different clusters have low similarity to each other}}. However, existing methods fail to well consider this criterion, \emph{i.e.}, \emph{\textbf{neglecting the existence of false-negative pairs}} .

In this article, by leveraging pseudo clustering labels $\overrightarrow{\textbf{L}}$, we can easily get the samples' index of different clusters.  As shown in \textbf{Figure~\ref{figill} (b)}, we aim to maximize the similarities of intra-cluster nodes, \emph{i.e.}, positive pairs, while minimizing the similarities of inter-cluster nodes, \emph{i.e.}, negative pairs. To this end, we first map the node representations $\textbf{Z}^{\textrm{(1)}}$ and $\textbf{Z}^{\textrm{(2)}}$ to obtain enhanced node representations $\textbf{M}^{\textrm{(1)}}$ and $\textbf{M}^{\textrm{(2)}}$ via a shared two-layer fully connected network with parameter $\bm{\phi}$, which also help to form and preserve more information in $\textbf{Z}^{\textrm{(1)}}$ and $\textbf{Z}^{\textrm{(2)}}$, where $\textbf{M}^{\textrm{(\emph{v})}} \in \mathbb{R}^{\emph{\textrm{N}}\times \emph{\textrm{d}}_\textrm{2}}$, $\emph{\textrm{d}}_\textrm{2}$ is the dimension of new node representation. After that, for the $\emph{\textrm{i}}$-th node, we propose a new self-supervised contrastive loss function, which is defined as
\begin{equation}\label{6}
\begin{aligned}
\mathcal{L}_\textrm{\emph{i}}\textrm{ = }\frac{{\textrm{-1}}}{{| {\Delta_\emph{\textrm{\emph{i}}}} |}}\sum_{{\emph{\textrm{t}} \in  \Delta_\emph{\textrm{i}}}}^{}\sum_{\alpha\textrm{,}\beta\textrm{=}\textrm{1}}^{\textrm{2}}{\textrm{{log}}}\frac{\textrm{e}^{\textrm{(}\circledS\textrm{(}{\textbf{m}}_\textrm{\emph{i}}^{\textrm{(}\alpha\textrm{)}}\textrm{, }{{\textbf{m}}}_\emph{\textrm{t}}^{\textrm{(}\beta\textrm{)}}\textrm{)}\textrm{/}\tau_\textrm{2}\textrm{)}}}{\sum\limits_{\alpha\textrm{'}\textrm{,}\beta\textrm{'}\textrm{=1}}^{\textrm{2}}\sum\limits_{q \in  \nabla_\emph{\textrm{i}}} {{{\textrm{{e}}}^{( {\circledS({{\textbf{m}}}_\emph{\textrm{i}}^{\textrm{(}\alpha\textrm{'}\textrm{)}}\textrm{, }{\bf{m}}_{\textrm{\emph{q}}}^{(\beta\textrm{'})}\textrm{)}\textrm{/}\tau_2 } \textrm{)}}}}}\textrm{,}
\end{aligned}
\end{equation}
where $\tau_\textrm{2}$ is the temperature parameter, $\textbf{m}^{\textrm{(\emph{v})}}_i$ represents the $\emph{\textrm{i}}$-th row of node representation $\textbf{M}^{\textrm{(\emph{v})}}$. $\Delta_\textrm{\emph{i}}$ represents the set of nodes that belong to the same cluster as the $\textrm{\emph{i}}$-th node, and $| {\Delta\textrm{(\emph{i})}} |$ is its cardinality, which can be obtained from the pseudo clustering assignment matrix $\overrightarrow{\textbf{L}}$. $\nabla_\emph{\textrm{i}}$ is the set of indices of all nodes except the $\textrm{\emph{i}}$-th node.

Then, taking all nodes into account, the self-supervised contrastive loss is
\begin{equation}\label{7}
\begin{aligned}
\mathcal{L}_{\textrm{SGC}} \textrm{ = } \underset{\bm{\Omega}\textrm{, }\bm{\phi}}{\textrm{min}}\sum_{\textrm{\emph{i}=1}}^{\textrm{\emph{N}}}\mathcal{L}_\emph{\textrm{i}}\textrm{.}
\end{aligned}
\end{equation}

\subsection{Contrastive Clustering Module}
How to obtain the clustering labels is crucial for downstream clustering task. Most existing methods directly implement classical clustering algorithms, \emph{e.g.}, \emph{K}-Means or spectral clustering, on the learned node representation to get clustering results. However, such strategy executes the node representation and clustering in two separated steps, which limits clustering performance. To this end, we build a clustering network to directly obtain the clustering labels. Specifically, as shown in Figure~\ref{fig1}, the clustering network is applied to transform the pattern structures of $\textbf{Z}^{\textrm{(1)}}$ and $\textbf{Z}^{\textrm{(1)}}$ into probability distribution of clustering labels $\widehat{\textbf{L}}^{\textrm{(1)}}$ and $\widehat{\textbf{L}}^{\textrm{(2)}}$.

To share the parameters across augmentations, we execute $\widehat{\textbf{L}}^{\textrm{(1)}}$ and $\widehat{\textbf{L}}^{\textrm{(2)}}$ through a shared two-layer fully connected network with parameter $\bm{\psi}$. Under this setting, we can ensure $\widehat{\textbf{L}}^{\textrm{(1)}}$ and $\widehat{\textbf{L}}^{\textrm{(2)}}$ own the same coding scheme. Thus, $\widehat{\textbf{L}}^{\textrm{(1)}} \in \mathbb{R}^{\emph{\textrm{N}}\times \emph{\textrm{K}}}$ is the output of clustering network under the 1-st augmented attributed graph view, and $\widehat{\textbf{L}}^{\textrm{(2)}}$ for the 2-nd augmented attributed graph view, where $\emph{\textrm{K}}$ is the number of clusters, $\hat{\bm{\ell}}^{\textrm{(1)}}_{\emph{\textrm{i}}\textrm{, }\emph{\textrm{k}}}$ represents the probability that assigning the $\emph{\textrm{i}}$-th node to the $\emph{k}$-the cluster $\textbf{C}_\emph{\textrm{k}}$.

For the obtained assignment matrices $\widehat{\textbf{L}}^{\textrm{(1)}}$ and $\widehat{\textbf{L}}^{\textrm{(2)}}$, in the column direction, each column $\hat{\bm{\ell}}^{\textrm{(1)}}_{\textrm{, }\textrm{\emph{k}}}$ of $\widehat{\textbf{L}}^{\textrm{(1)}}$ is the representation of the $\emph{\textrm{k}}$-th cluster. Thus, \emph{\textbf{we should push closer the cluster representation of the same class, and also push far away the cluster representation of different class}}. That is to say, for the $\emph{\textrm{k}}$-th cluster in each augmented attributed graph view, there is only one positive pair $\textrm{(}\hat{\bm{\ell}}^{\textrm{(\emph{v})}}_{\textrm{, }\emph{\textrm{k}}}\textrm{, }\hat{\bm{\ell}}^{\textrm{(\emph{v})}}_{\textrm{, }\textrm{\emph{k}}}\textrm{)}$, and $\textrm{2\emph{K}-2}$ negative pairs. To this end, motivated by the great success of contrastive learning~\cite{ChenK0H20}, we leverage the contrastive loss function to implement this constraint. Thus, for the $\textrm{\emph{k}}$-th cluster in the 1-st augmentation, we have
\begin{equation}\label{3}
\begin{aligned}
\mathcal{L}\textrm{(}\hat{\bm{\ell}}^{\textrm{(1)}}_{\textrm{, \emph{k}}}\textrm{,}\hat{\bm{\ell}}^{\textrm{(2)}}_{\textrm{, }\textrm{\emph{k}}}\textrm{)}\textrm{=-}\textrm{log}\frac{\textrm{e}^{\textrm{(}\circledS(\hat{\bm{\ell}}^{\textrm{(1)}}_{\textrm{, }\emph{\textrm{k}}}\textrm{,}\hat{\bm{\ell}}^{\textrm{(2)}}_{\textrm{, }\textrm{\emph{k}}}\textrm{)}\textrm{/}\tau_\textrm{1}\textrm{)}}}{\underset{\textrm{inter-view pairs}}{\underbrace{{\sum_{\textrm{\emph{j}}=1}^{\emph{\textrm{K}}}\textrm{e}^{\textrm{(}\circledS(\hat{\bm{\ell}}^{\textrm{(1)}}_{\textrm{, }\emph{\textrm{k}}}\textrm{,}\hat{\bm{\ell}}^{\textrm{(1)}}_{\textrm{, }\textrm{\emph{j}}}\textrm{)}\textrm{/}\tau_\textrm{1}\textrm{)}}}}}\textrm{+}\underset{\textrm{intra-view pairs}}{\underbrace{{\sum_{j=1}^{\emph{\textrm{K}}}\textrm{e}^{\textrm{(}\circledS\textrm{(}\hat{\bm{\ell}}^{\textrm{(1)}}_{\textrm{, }\emph{\textrm{k}}}\textrm{,}\hat{\bm{\ell}}^{\textrm{(2)}}_{\textrm{, }\textrm{\emph{j}}}\textrm{)}\textrm{/}\tau_\textrm{1}\textrm{)}}}}}}\textrm{,}
\end{aligned}
\end{equation}
where $\tau_\textrm{1}$ is parameter to control the softness. Given two vectors $\textbf{f}$ and $\textbf{s}$, $\circledS\textrm{(}{\textbf{f}\textrm{, }\textbf{s}}\textrm{)}$ is the cosine similarity between them. In this article, we use the function $\circledS\textrm{(}{\cdot\textrm{, }\cdot}\textrm{)}$ to measure the similarity of node pairs. Then, taking all positive pairs into account, the contrastive clustering loss $\mathcal{L}_\textrm{{CC}}$ is defined as
\begin{equation}\label{4}
\begin{aligned}
\mathcal{L}_\textrm{{CC}}\textrm{ = }\underset{\bm{\Omega}\textrm{, }\bm{\psi}}{\textrm{min}}\frac{\textrm{1}}{\textrm{2\emph{K}}}\sum_{\textrm{\emph{k}=1}}^{\emph{\textrm{K}}}\left[\mathcal{L}\textrm{(}\hat{\bm{\ell}}^{\textrm{(1)}}_{\textrm{, \emph{k}}}\textrm{,}\hat{\bm{\ell}}^{\textrm{(2)}}_{\textrm{, }\textrm{\emph{k}}}\textrm{)}\textrm{+}\mathcal{L}\textrm{(}\hat{\bm{\ell}}^{\textrm{(2)}}_{\textrm{, \emph{k}}}\textrm{,}\hat{\bm{\ell}}^{\textrm{(1)}}_{\textrm{, }\textrm{\emph{k}}}\textrm{)}
\right]\textrm{,}
\end{aligned}
\end{equation}

Moreover, to avoid trivial solution, \emph{i.e.}, making sure that all nodes could be evenly assigned into all clusters, similar to~\cite{Li0LPZ021,MaoYGY21}, we herein introduce a clustering regularizer $\mathcal{R}$, which is defined as
\begin{equation}\label{5}
\begin{aligned}
\mathcal{R}\textrm{ = }\underset{\bm{\Omega}\textrm{, }\bm{\psi}}{\textrm{min}}\textrm{-}\sum_\textrm{{\emph{k}=1}}^{\emph{\textrm{K}}}\textrm{[}\rho\textrm{(}\hat{\bm{\ell}}^{\textrm{(1)}}_{\textrm{, }\emph{\textrm{k}}}\textrm{)}\textrm{log}\textrm{(}\hat{\bm{\ell}}^{\textrm{(1)}}_{\textrm{, }\textrm{\emph{k}}}\textrm{)}
\textrm{ + }\rho\textrm{(}\hat{\bm{\ell}}^{\textrm{(2)}}_{\textrm{, }\emph{\textrm{k}}}\textrm{)}\textrm{log}\textrm{(}\hat{\bm{\ell}}^{\textrm{(2)}}_{\textrm{, }\textrm{\emph{k}}}\textrm{)}
\textrm{)}\textrm{]}\textrm{,}
\end{aligned}
\end{equation}
where $\rho\textrm{(}\hat{\bm{\ell}}^{\textrm{(\emph{v})}}_{\textrm{, }\emph{\textrm{k}}}\textrm{)}  = \sum_{\textrm{\emph{i}=1}}^{\emph{\textrm{N}}}\frac{\hat{\bm{\ell}}^{\textrm{(\emph{v})}}_{\emph{\textrm{i}}\textrm{, }\emph{\textrm{k}}}}{\|\widehat{\textbf{L}}^{\textrm{(\emph{v})}}\|_\textrm{1}}$.

In the proposed SCAGC training process, when we take the un-augmented attributed graph $(\textbf{X}\textrm{, }\textbf{G})$ as the input of SCAGC, then we can get the clustering assignment matrix $\overrightarrow{\textbf{L}}$ by discretizing the continuous output probability $\widehat{\textbf{L}}$.

\begin{remark}
\textbf{Solving out-of-sample nodes}. For OOS nodes $(\textbf{X}_{\textrm{new}}\textrm{, }\textbf{G}_{\textrm{new}})$, SCAGC can directly take $(\textbf{X}_{\textrm{new}}\textrm{, }\textbf{G}_{\textrm{new}})$ as input to calculate the clustering assignment matrix. While existing GRL and GCRL based methods is inefficient in OOS nodes $(\textbf{X}_{\textrm{new}}\textrm{, }\textbf{G}_{\textrm{new}})$, which require training the whole attributed graph, i.e., \{$\textrm{(}\textbf{X}\textrm{; }\textbf{X}_{\textrm{new}}\textrm{)}\textrm{, }\textrm{(}\textbf{G}\textrm{; }\textbf{G}_{\textrm{new}}\textrm{)}$\}.
\end{remark}

\begin{algorithm}[!t]
  \caption{Procedure for training SCAGC}
  \label{A1}
  \LinesNumbered
  \KwIn{Attributed graph with node attribute matrix $\textbf{X}$ and adjacency matrix $\textbf{G}$, cluster number $\emph{\textrm{K}}$, hyper-parameters $\tau_1$, $\tau_2$, $\gamma$, learning rate and maximum number of iterations $\textrm{T}_\textrm{max}$.}
  \KwOut{Clustering label $\overrightarrow{\textbf{L}}$.}
  \textbf{Initialization}: initialize the parameters ${\bm{\Omega}\textrm{, }\bm{\phi}\textrm{, }\bm{\psi}}$ of each component, the clustering assignment matrix $\overrightarrow{\textbf{L}}$ by inputting raw attributed graph $(\textbf{X}\textrm{, }\textbf{G})$\;
  \tcp{Training SCAGC}
  \For{$\textrm{\emph{T}} = 1: \textrm{\emph{T}}_\textrm{max} $}
  {Sample two stochastic graph augmentation schemes $\emph{\textbf{A}}^{\textrm{(1)}}\thicksim \bm{\mathcal{A}}$ and $\emph{\textbf{A}}^{\textrm{(2)}}\thicksim \bm{\mathcal{A}}$\;
  Construct the augmented attributed graph views: where $\textbf{X}^{\textrm{(1)}}=\textbf{\emph{A}}^{\textrm{(1)}}(\textbf{X})$,  $\textbf{G}^{\textrm{(1)}}=\textbf{\emph{A}}^{\textrm{(1)}}(\textbf{G})$, $\textbf{X}^{\textrm{\textrm{(2)}}}=\textbf{\emph{A}}^{\textrm{\textrm{(2)}}}(\textbf{X})$, and $\textbf{G}^{(2)}=\textbf{\emph{A}}^{\textrm{(2)}}(\textbf{G})$\;
  Obtain variables $\textbf{Z}^{\textrm{(1)}}$, $\textbf{Z}^{\textrm{(2)}}$, $\textbf{M}^{\textrm{(1)}}$, $\textbf{M}^{\textrm{(2)}}$, $\widehat{\textbf{L}}^{\textrm{(1)}}$ and $\widehat{\textbf{L}}^{\textrm{(2)}}$ by forward propagation\;
  Calculate the overall objective with Eq. (\ref{8}) and pseudo clustering label $\overrightarrow{\textbf{L}}$\;
  Update network parameters ${\bm{\Omega}\textrm{, }\bm{\phi}\textrm{, }\bm{\psi}}$ via stochastic gradient ascent to minimize Eq. (\ref{8})\;
  \tcp{Update pseudo clustering label}
  \If{T \% 5 ==0}
  {Update the clustering assignment matrix $\overrightarrow{\textbf{L}}$ by mapping raw attributed graph $(\textbf{X}\textrm{, }\textbf{G})$\;}
  }
  \tcp{Obtain clustering results}
  Obtain the clustering assignment matrix $\overrightarrow{\textbf{L}}$ by mapping raw attributed graph $(\textbf{X}\textrm{, }\textbf{G})$\;
  \textbf{return:} Clustering label matrix $\overrightarrow{\textbf{L}}$.
\end{algorithm}

\subsection{Optimization}
Finally, we integrate the aforementioned three sub-modules into an end-to end optimization framework, the overall objective function of SCAGC can be formulated as
\begin{equation}\label{8}
\begin{aligned}
\mathcal{L}_{\textrm{Total}}\textrm{ = }\underset{\bm{\Omega}\textrm{, }\bm{\phi}\textrm{, }\bm{\psi}}{\textrm{min}}\mathcal{L}_{\textrm{SGC}}\textrm{ + } \mathcal{L}_{\textrm{CC}}\textrm{ + }\gamma\mathcal{R}\textrm{,}
\end{aligned}
\end{equation}
where $\gamma$ is a trade-off parameter. By optimizing Eq. (\ref{8}), some nodes with correct labels will propagate useful information for graph representation learning, where the latter is used in turn to conduct the sub-sequent clustering. By this strategy, the node clustering and graph representation learning are seamlessly connected, with the aim to achieve better clustering results. We employ Adam optimizer~\cite{KingmaB14} with learning rate $\eta$ to optimize the proposed SCAGC, \emph{i.e.}, Eq. (\ref{8}). Algorithm~\ref{A1} presents the pseudo-code of optimizing the proposed SCAGC.

\begin{table*}[!t]
\begin{center}
\caption{Statistics of the real-world evaluation datasets.}\label{Dataset}
\vspace{-2mm}
\resizebox{1.0\columnwidth}{!}{
\begin{tabular}{c|c|c|c|c|c|c}
\toprule
Dataset&\# Nodes&\# Attribute dimension& \# Edges & \# Classes & Type & Scale\\
\midrule
ACM~\cite{TangZYLZS08}&3, 025&1, 870&29, 281&3&Paper relationship& Small\\
DBLP~\cite{PanWZZW16}&4, 057&334&5, 000, 495&4&Author relationship& Small\\
Amazon-Photo~\cite{DataN}&7, 650&745&119, 081&8&Commodity purchase relationship& Medium\\
Amazon-Computers~\cite{DataN}&13, 752& 767&245, 861&10&Commodity purchase relationship& Large\\
\bottomrule
\end{tabular}}
\end{center}
\vspace{-4mm}
\end{table*}
\section{Experiments}
\subsection{Experiment Setup}
\subsubsection{Benchmark Datasets} In this article, we use four real-world attributed graph datasets from different domains, \emph{e.g.}, academic network, shopping network, to evaluate the effectiveness of the proposed SCAGC, including ACM\footnote{\url{http://dl.acm.org}}, DBLP~\footnote{\url{https://dblp.uni-trier.de/}}, Amazon-Photo\footnote{\url{https://github.com/shchur/gnn-benchmark/raw/master/data/npz/amazon_electronics_photo.npz}} and Amazon-Computers\footnote{\url{https://github.com/shchur/gnn-benchmark/raw/master/data/npz/amazon_electronics_computers.npz}}. Table~\ref{Dataset} presents detailed statistics of these datasets.

\subsubsection{Baseline Methods} We compare clustering performance of the proposed SCAGC with 11 state-of-the-art node clustering methods, including the following three categories:
\begin{enumerate}
  \item \textbf{Classical clustering methods}: \emph{K}-means, and spectral clustering (SC);
  \item \textbf{Graph embedding clustering methods}: GAE~\cite{KipfW16a}, VGAE~\cite{KipfW16a}, ARGA~\cite{PanHFLJZ20}, ARVGA~\cite{PanHFLJZ20}, DAEGC~\cite{WangPHLJZ19}, SDCN~\cite{Bo0SZL020}, and DFCN~\cite{TuZL0CZC21}.
  \item \textbf{GCRL based methods}: GraphCL~\cite{YouCSCWS20} and GCA~\cite{XYLWW21}.
\end{enumerate}

For the first category, \emph{K}-means takes raw node attribute as input, and SC takes raw topological graph structure as input. As for the second and third categories, they take raw node attribute and topological graph structure as input. For GAE, VGAE, ARGA, ARVGA, SDCN, DFCN, GraphCL and GCA, the clustering assignment matrix is obtained by running \emph{K}-means on the extracted node representation.

\subsubsection{Evaluation Metrics} Similar to~\cite{Bo0SZL020,TuZL0CZC21}, we leverage four commonly used metrics to evaluate the efficiency of all methods, \emph{i.e.}, accuracy (ACC), normalized mutual information (NMI), average rand index (ARI), and macro F1-score (F1). For these metrics, the higher the value, the better the performance.

\begin{table*}[!t]
	\centering
\caption{The clustering results on ACM and DBLP benchmarks. The best results in all methods and all baselines are represented by \textbf{bold} value and \underline{underline} value, respectively.}\label{T1}
    \resizebox{1.0\columnwidth}{!}{
\begin{tabular}{c|cccc|cccc}
	\toprule		
Dataset&\multicolumn{4}{c|}{ACM}&\multicolumn{4}{c}{DBLP}\\
\midrule
Metric&ACC ($\uparrow$)&NMI ($\uparrow$)&$\text{F}_1$ ($\uparrow$)&ARI ($\uparrow$)&ACC ($\uparrow$)&NMI ($\uparrow$)&$\text{F}_1$ ($\uparrow$)&ARI ($\uparrow$)\\
\midrule
\emph{K}-Means&67.26 $\pm$ 0.75&31.91 $\pm$ 0.35&54.47 $\pm$ 0.32&30.76 $\pm$ 0.62& 39.08 $\pm$ 0.36&10.11 $\pm$ 0.21&38.01 $\pm$ 0.37&7.28 $\pm$ 0.29\\
SC&36.80 $\pm$ 0.00&0.75 $\pm$ 0.00&42.63 $\pm$ 0.00&0.58 $\pm$ 0.00&29.57 $\pm$ 0.01&0.08 $\pm$ 0.00&40.86 $\pm$ 0.00&0.70 $\pm$ 0.00\\
\midrule
GAE (\emph{NeurIPS}' 16)&82.47 $\pm$ 0.92&50.29 $\pm$ 1.86&82.65 $\pm$ 0.89	&54.59 $\pm$ 1.99&59.25 $\pm$ 0.40&26.37 $\pm$ 0.29&59.84 $\pm$ 0.32&20.95 $\pm$ 0.43\\
VGAE (\emph{NeurIPS}' 16)&82.85 $\pm$ 0.63&50.22 $\pm$ 1.24&82.85 $\pm$ 0.62& 55.56 $\pm$ 1.15&62.22 $\pm$ 0.83&26.62 $\pm$ 1.37&60.70 $\pm$ 0.85&25.08 $\pm$ 1.23\\
ARGA (\emph{IEEE TC}' 20)&86.85 $\pm$ 0.64&58.05 $\pm$ 1.53&86.84 $\pm$ 0.60&64.77 $\pm$ 1.53&64.60 $\pm$ 0.95& 28.65 $\pm$ 0.63&64.49 $\pm$ 0.63&27.44 $\pm$ 1.27\\
ARVGA (\emph{IEEE TC}' 20)&84.84 $\pm$ 0.36&	52.89 $\pm$ 0.84&84.86 $\pm$ 0.35&59.67 $\pm$ 0.85&64.10 $\pm$ 0.96&31.01 $\pm$ 0.89	&64.36 $\pm$ 1.01&25.69 $\pm$ 1.51\\
DAEGC (\emph{IJCAI}' 19)&87.18 $\pm$ 0.05&59.32 $\pm$ 0.12&87.27 $\pm$ 0.05& 65.46 $\pm$ 0.12&\underline{75.87 $\pm$ 0.46}&42.45 $\pm$ 0.58&\underline{75.41 $\pm$ 0.45}&\underline{46.80 $\pm$ 0.87}\\
SDCN (\emph{WWW}' 20)&89.44 $\pm$ 0.26&65.89 $\pm$ 0.95&89.40 $\pm$ 0.28& 71.47 $\pm$ 0.67&71.91 $\pm$ 0.57&37.80 $\pm$ 1.06&71.21 $\pm$ 0.73&40.45 $\pm$ 1.18\\
DFCN (\emph{AAAI}' 21)&90.15 $\pm$ 0.05&67.98 $\pm$ 0.18&\underline{90.14 $\pm$ 0.05}&73.25 $\pm$ 0.14&75.42 $\pm$ 0.82&43.20 $\pm$ 0.74	&75.31 $\pm$ 0.71&45.07 $\pm$ 1.91\\
\midrule
GraphCL (\emph{NeurIPS}' 20)& \underline{90.18 $\pm$ 0.04} &\underline{68.24 $\pm$ 0.12} &90.04 $\pm$ 0.05 &\underline{73.38 $\pm$ 0.09} &74.90 $\pm$ 0.10 &\underline{45.14 $\pm$ 0.14} &74.51 $\pm$ 0.10 &45.86 $\pm$ 0.19\\
GCA (\emph{WWW}' 21)& 88.95 $\pm$ 0.26&65.33 $\pm$ 0.56&89.07 $\pm$ 0.26&69.82 $\pm$ 0.67&73.90 $\pm$ 0.48&41.35 $\pm$ 0.79&72.91 $\pm$ 0.76&43.65 $\pm$ 0.65\\
\midrule
SCAGC&\textbf{91.83} $\bm{\pm}$\textbf{ 0.03}&\textbf{71.28} $\bm{\pm}$ \textbf{0.06}&\textbf{91.84} $\bm{\pm}$ \textbf{0.03}&\textbf{77.29} $\bm{\pm}$ \textbf{0.07}&\textbf{79.42} $\bm{\pm}$ \textbf{0.02}&\textbf{49.05} $\bm{\pm}$ \textbf{0.02}&\textbf{78.88} $\pm$ \textbf{0.02}&\textbf{54.04} $\bm{\pm}$ \textbf{0.03}\\
\bottomrule
	\end{tabular}}
\end{table*}

\subsubsection{Implementation Details} The proposed SCAGC and the baseline methods are implemented on a Windows 10 machine with an Intel (R) Xeon (R) Gold 6230 CPU and dual NVIDIA Tesla P100-PCIE GPUs. The deep learning environment consists of PyTorch 1.6.0 platform, PyTorch Geometric 1.6.1 platform, and TensorFlow 1.13.1. To ensure the availability of the initial pseudo clustering assignment matrix $\overrightarrow{\textbf{L}}$, we pre-train the shared graph convolutional encoder and graph contrastive representation learning module via a classic contrastive learning loss.

The hyper-parameters of the proposed methods on each datasets are reported in \textbf{supplementary material}. In this article, we use the adaptive graph augmentation functions proposed by~\cite{XYLWW21} to augment node attribute and topological structure. Notably, the degree centrality is used as the node centrality function to generate different topology graph views. The output size of shared graph convolutional encoder is set to 256, the output size of graph contrastive representation learning sub-network is set to 128, and the output size of contrastive clustering sub-network is set to be equal to the number of clusters $\emph{\textrm{K}}$.

For all baseline methods, we follow the hyper-parameter settings as reported in their articles and run their released code to obtain the clustering results. To avoid the randomness of the clustering results, we repeat each experiment of SCAGC and baseline methods for 10 times and report their average values and the corresponding standard deviations.

\begin{table*}[!t]
	\centering
\caption{The clustering results on Amazon-Photo and Amazon-Computers benchmarks. The best results in all methods and all baselines are represented by \textbf{bold} value and \underline{underline} value, respectively.}\label{T2}
    \resizebox{1.0\columnwidth}{!}{
\begin{tabular}{c|cccc|cccc}
	\toprule		
Dataset&\multicolumn{4}{c|}{Amazon-Photo}&\multicolumn{4}{c}{Amazon-Computers}\\
\midrule
Metric&ACC ($\uparrow$)&NMI ($\uparrow$)&$\text{F}_1$ ($\uparrow$)&ARI ($\uparrow$)&ACC ($\uparrow$)&NMI ($\uparrow$)&$\text{F}_1$ ($\uparrow$)&ARI ($\uparrow$)\\
\midrule
\emph{K}-Means&36.53 $\pm$ 4.11&19.31 $\pm$ 3.75& 32.63 $\pm$ 1.90&12.61 $\pm$ 3.54
& 36.44 $\pm$ 2.64&16.64 $\pm$ 4.59&28.08 $\pm$ 1.44&2.71 $\pm$ 1.98\\
SC &25.58 $\pm$ 0.02&0.60 $\pm$ 0.02&5.50 $\pm$ 0.00&0.03 $\pm$ 0.00
&36.47 $\pm$ 0.01 & 0.37 $\pm$ 0.02  & 5.81 $\pm$ 0.00  & 0.59 $\pm$ 0.00  \\
\midrule
GAE (\emph{NeurIPS}' 16)&42.03 $\pm$ 0.54&31.87 $\pm$ 0.51&34.01 $\pm$ 0.42	&19.31 $\pm$ 0.53&43.14 $\pm$ 1.74 & 35.47 $\pm$ 1.58 & 27.06 $\pm$ 2.63 & 19.61 $\pm$ 1.85 \\
VGAE (\emph{NeurIPS}' 16)&40.67 $\pm$ 0.92&31.46 $\pm$ 2.03&38.01 $\pm$ 2.67& 15.70 $\pm$ 1.18&42.44 $\pm$ 0.16 & 37.62 $\pm$ 0.23 & 24.94 $\pm$ 0.14 & 22.16 $\pm$ 0.35 \\
ARGA (\emph{IEEE TC}' 20)&57.79 $\pm$ 2.26&48.01 $\pm$ 1.65&52.56 $\pm$ 2.68&34.44 $\pm$ 1.58&45.67 $\pm$ 0.37 & 37.21 $\pm$ 0.92 & 40.02 $\pm$ 1.29 & 26.28 $\pm$ 1.02 \\
ARVGA (\emph{IEEE TC}' 20)&47.89 $\pm$ 1.36&	41.37 $\pm$ 1.39&42.96 $\pm$ 1.46&27.72 $\pm$ 1.06&47.16 $\pm$ 0.26 & 38.84 $\pm$ 0.96 & \underline{41.51 $\pm$ 0.83} & 27.27 $\pm$ 0.84 \\
DAEGC (\emph{IJCAI}' 19)&60.14 $\pm$ 0.93&58.03 $\pm$ 1.25&52.37 $\pm$ 2.39& 43.55 $\pm$ 1.76&49.26 $\pm$ 0.49 & 39.28 $\pm$ 4.97 & 33.71 $\pm$ 5.76 & 35.29 $\pm$ 1.97 \\
SDCN (\emph{WWW}' 20)&71.43 $\pm$ 0.31&64.13 $\pm$ 0.10&68.74 $\pm$ 0.22& 51.17 $\pm$ 0.13&54.12 $\pm$ 1.13 & 39.90 $\pm$ 1.51 & 28.84 $\pm$ 4.20 & 31.59 $\pm$ 1.08 \\

DFCN (\emph{AAAI}' 21)&\underline{73.43 $\pm$ 0.61}&\underline{64.74 $\pm$ 1.04}&\underline{69.96 $\pm$ 0.49}&\underline{52.39 $\pm$ 1.01}&\underline{56.24 $\pm$ 0.16} & 41.83 $\pm$ 0.40 & 33.39 $\pm$ 1.11 & 33.02 $\pm$ 0.39 \\
\midrule
GraphCL (\emph{NeurIPS}' 20)& 66.61 $\pm$ 0.56 &57.35 $\pm$ 0.32 &58.52 $\pm$ 0.55 &45.13 $\pm$ 0.44 &50.22 $\pm$ 0.66 & 41.78 $\pm$ 2.44 & 32.89 $\pm$ 2.16 & \underline{36.94 $\pm$ 3.20}\\
GCA (\emph{WWW}' 21)& 71.17 $\pm$ 0.27&60.70 $\pm$ 0.41&64.12 $\pm$ 1.21&49.09 $\pm$ 0.62&54.92 $\pm$ 0.55&\underline{44.36 $\pm$ 0.86}&40.43 $\pm$ 0.45&35.61 $\pm$ 0.62\\
\midrule
SCAGC&\textbf{75.25} $\bm{\pm}$\textbf{0.10}&\textbf{67.18} $\bm{\pm}$ \textbf{0.13}&\textbf{72.77} $\bm{\pm}$ \textbf{0.16}&\textbf{56.86} $\bm{\pm}$ \textbf{0.23}&\textbf{58.43} $\bm{\pm}$ \textbf{0.12}&\textbf{49.92} $\bm{\pm}$ \textbf{0.08}&\textbf{43.14} $\pm$ \textbf{0.09}&\textbf{38.29} $\bm{\pm}$ \textbf{0.07}\\
\bottomrule
	\end{tabular}}
\end{table*}

\subsection{Node Clustering Performance}
Table~\ref{T1} and Table~\ref{T2} present the node clustering results of the proposed SCAGC and all baseline methods. From these results, we have the following observations:
\begin{enumerate}
  \item The proposed SCAGC and other GCN based methods (GAE, VGAE, ARGA, ARVGA, DAEGC, SDCN, DFCN, GraphCL, GCA) significantly and consistently outperforms \emph{K}-Means and SC. The reason may be that GCN based methods simultaneously explore the information embedded in node attribute and topological graph structure. In contrast, these classical clustering methods only use the node attribute or topological structure. Moreover, compared with classical clustering methods, GCN based methods uses a multi-layer nonlinear graph neural network as the feature extractor, then map input data into a new subspace to carry out downstream clustering. These results well demonstrate the effectiveness of GCN on processing attributed graph data.
  \item The proposed SCAGC achieves much better clustering results than some representative graph auto-encoder (GAE, VGAE, ARGA, ARVGA). This is because compared with traditional graph auto-encoder, SCAGC leverages graph augmentation scheme to generate useful attributed graph, and take the relationship between positive pair and negative pair into account. These strategies help to improve the quality of node representation.
  \item In some cases, the clustering performance of GCL based baselines, \ie, GraphCL and GCA, are inferior to clustering-directed, \ie, DAEGC, SDCN, DFCN and the proposed SCAGC. This is because SCAGC integrate the node clustering and representation into an end-to-end framework, which helps to better explore the cluster structure. In contrast, GraphCL and GCA execute the node representation and clustering in two separated steps, which limits their performances.
  \item The proposed SCAGC consistently outperforms all the state-of-the-art baselines on all four datasets. Particularly, SCAGC surpasses the closest competitor GCA by 5.95\% on ACM and 7.7\% on DBLP, in terms of NMI. These remarkable performance verify the clustering ability of SCAGC. And it demonstrates that contrastive clustering module and self-supervised graph contrastive representation learning module are effective at benefiting the node representation learning and clustering.
\end{enumerate}

\begin{figure}[!t]
\centering
\subfigure[ACM]{
\begin{minipage}[t]{0.47\linewidth}
\centering
\includegraphics[width=0.8\linewidth]{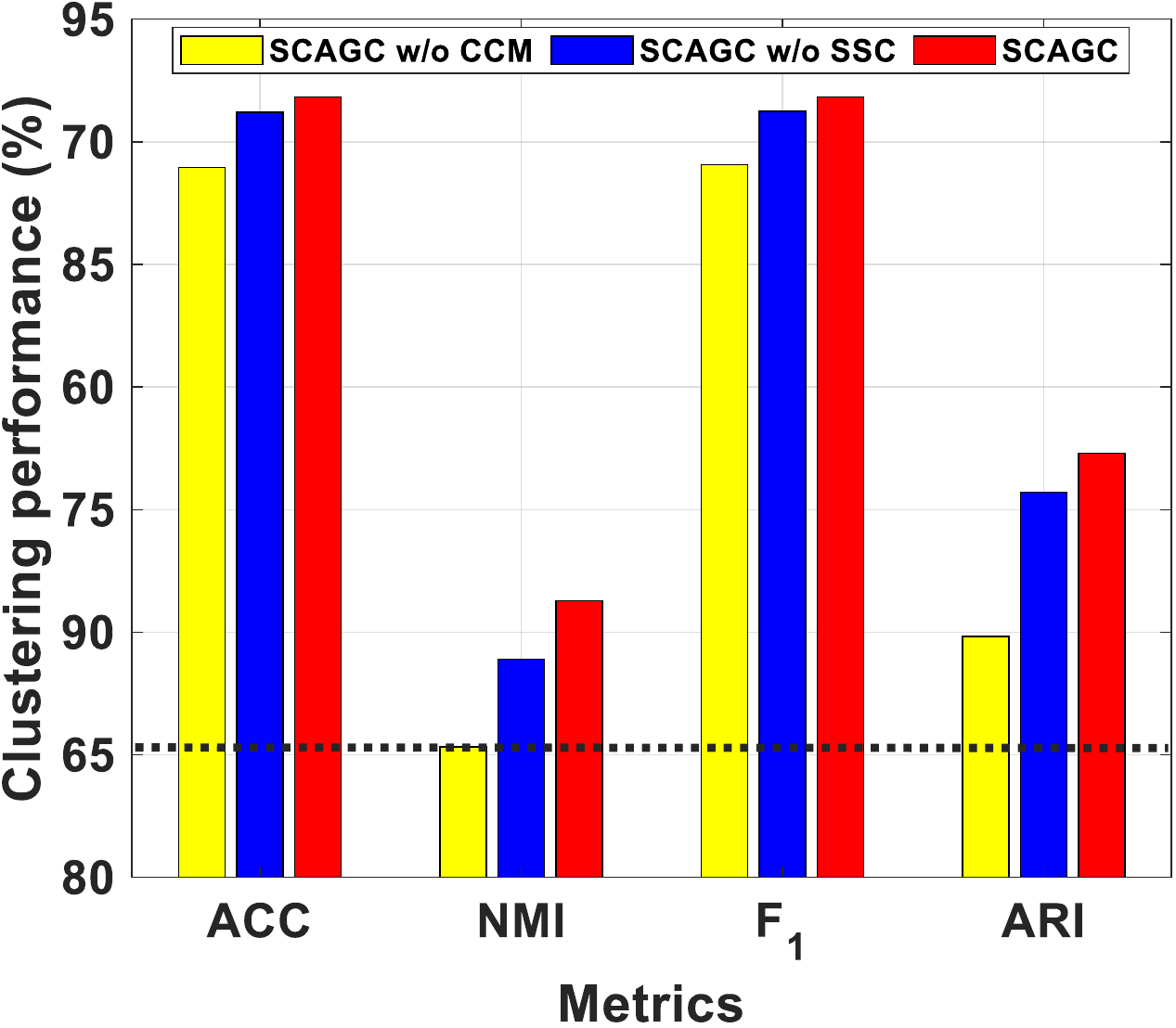}
\end{minipage}
}
\subfigure[DBLP]{
\begin{minipage}[t]{0.47\linewidth}
\centering
\includegraphics[width=0.8\linewidth]{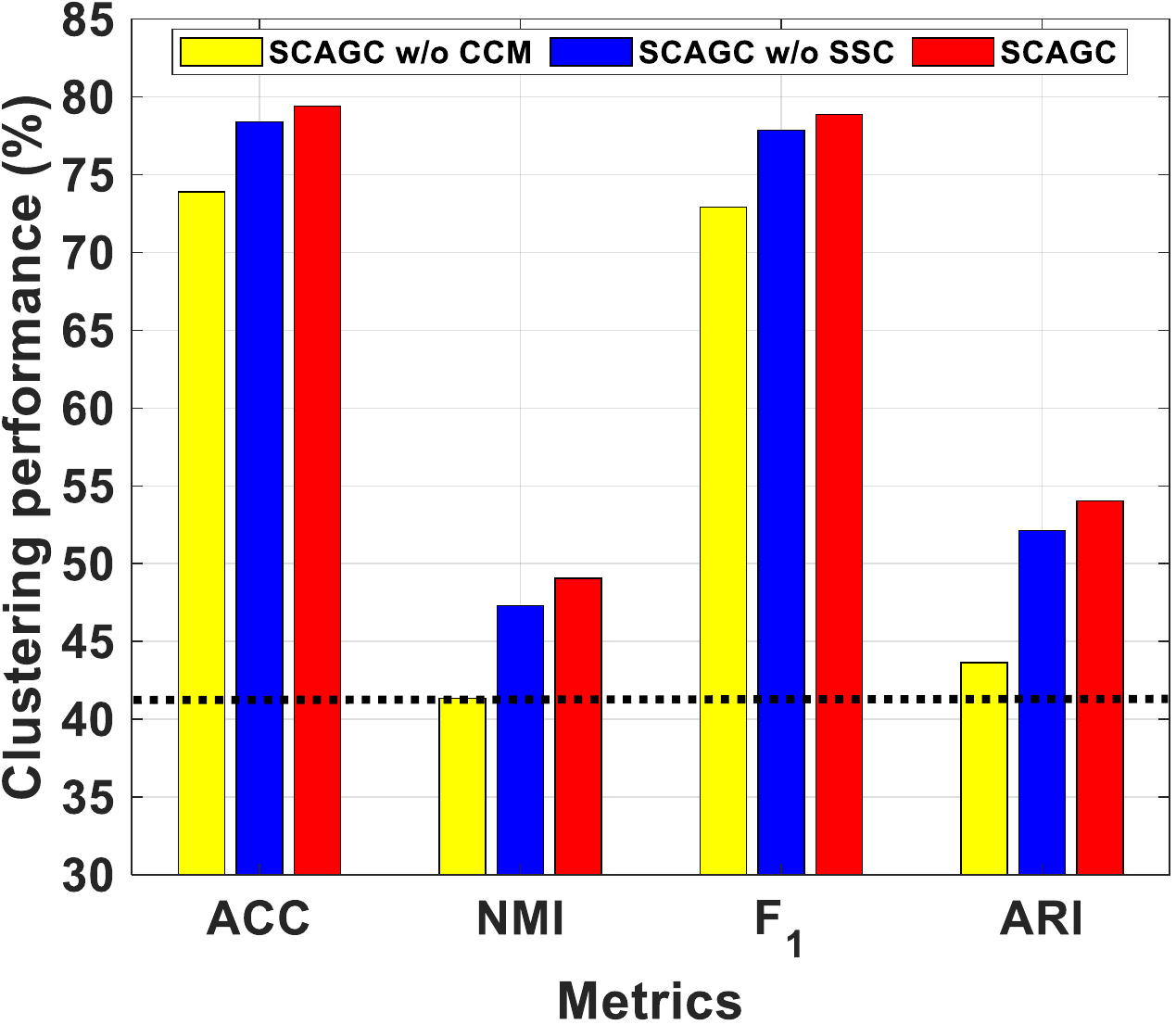}
\end{minipage}
}
\centering
\caption{Ablation Studies on ACM and DBLP datasets.}
\label{A-2}
\end{figure}

\begin{figure*}[!t]
	\centering
	\includegraphics[width=1.0\linewidth]{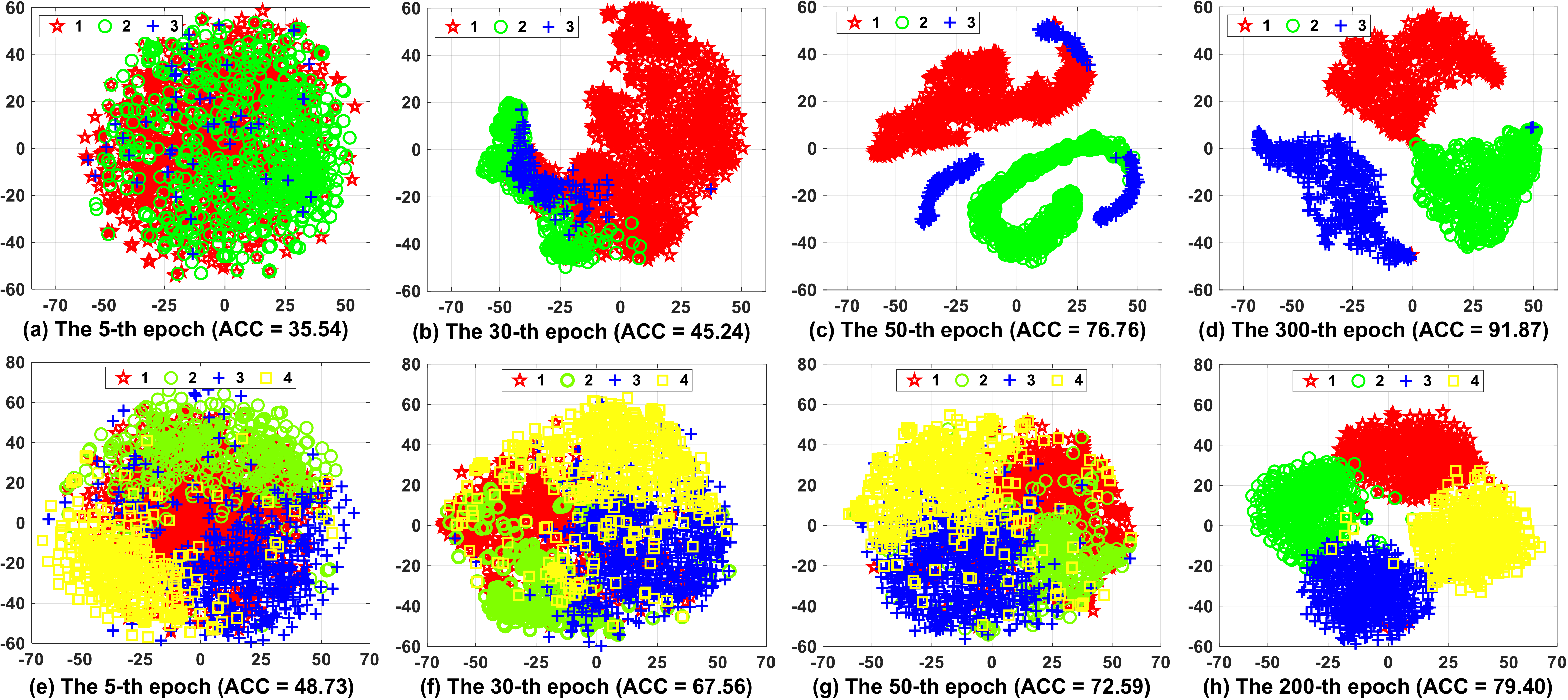}
	\caption{The t-SNE visualizations on the ACM (a-d) and IMDB (e-h) datasets with the increasing of the number of iteration.}
	\label{V-1}
\end{figure*}

\begin{figure}[!t]
	\centering
	\includegraphics[width=0.5\linewidth]{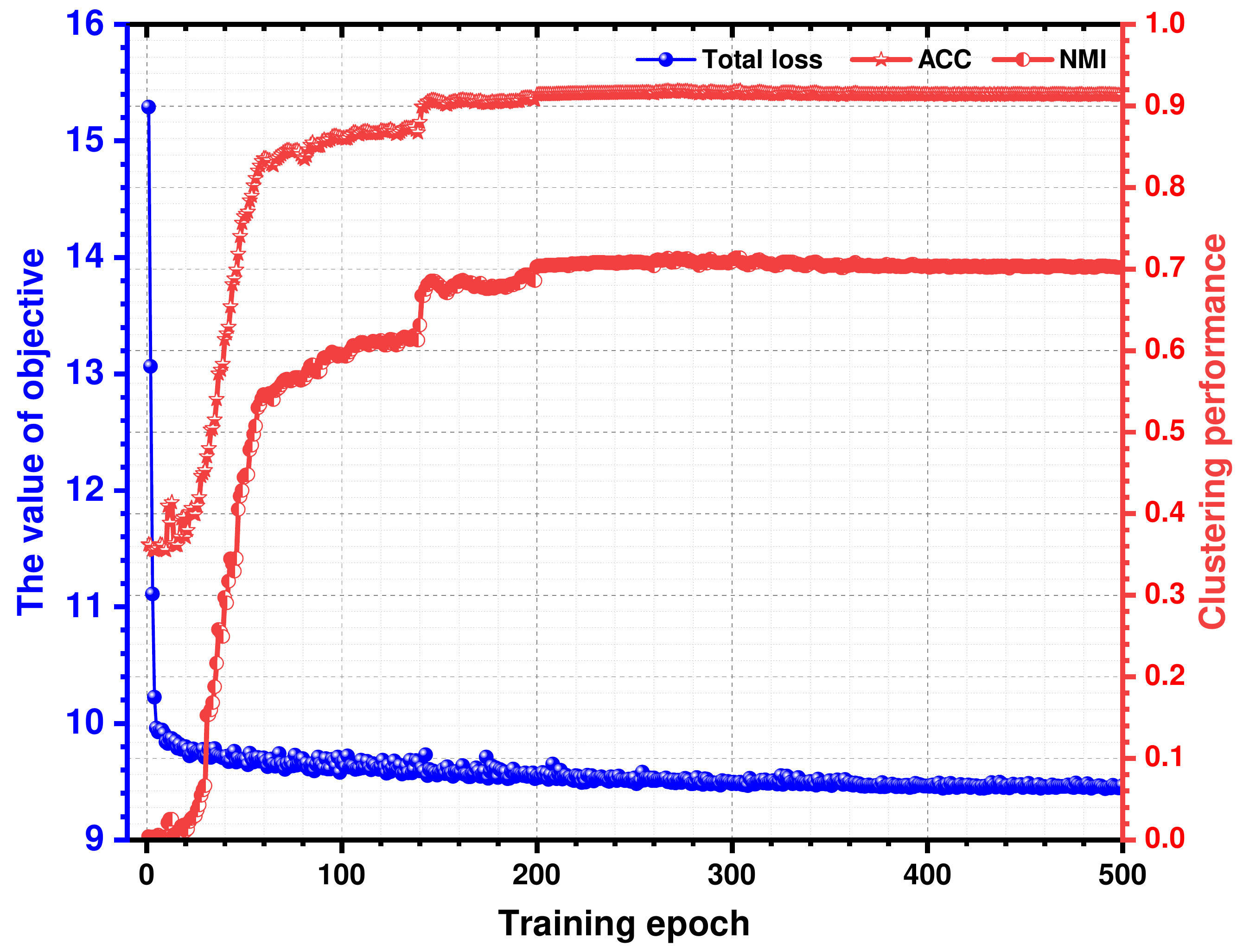}
	\caption{The convergence of SCAGC on ACM dataset.}
	\label{C-A}
\end{figure}

\subsection{Ablation Studies}
To better illustrate the effectiveness of different components in SCAGC, two ablation scenarios are implemented to further verify the effectiveness of contrastive clustering module, and the proposed self-supervised GCRL loss.

\subsubsection{Effect of Contrastive Clustering Module} To better illustrate the effectiveness of contrastive clustering module, we compare the clustering results of SCAGC and SCAGC without contrastive clustering module (\textbf{termed SCAGC w/o CCM}) on ACM and DBLP datasets. Note that, in this scenario, SCAGC w/o CCM is trained using traditional contrastive loss~\cite{ChenK0H20,XYLWW21}, \ie, SCAGC w/o CCM is clustering-agnostic. As shown in Figure~\ref{A-2} (a-b), the clustering performances of SCAGC (see the red bar) are substantially superior to SCAGC w/o CCM (see the yellow bar). This is because SCAGC can better extract node representation benefiting from contrastive clustering module. While in the absence of the specific clustering task, SCAGC w/o CCM fails to explore the cluster structure, resulting in the
quick drop of the performance of SCAGC.

\subsubsection{Importance of the Proposed Self-Supervised GCRL Loss} To this end, we compare the clustering performances of SCAGC and SCAGC without self-supervised GCRL loss (\textbf{termed SCAGC w/o SSC}) on ACM and DBLP datasets. Note that, in this scenario, SCAGC w/o CCM is trained by replacing the first term of Eq. (\ref{8}), \ie, Eq. (\ref{6}), to a standard contrastive loss~\cite{ChenK0H20,XYLWW21}. As reported in Figure~\ref{A-2} (a-b), SCAGC (see red bar) always achieves the best performance in terms of all four metrics. These results demonstrate that pseudo label supervision guides the GCRL, thus, leveraging clustering labels are promising methods for unsupervised clustering task.

\subsection{Model Discussion}

\subsubsection{Visualizations of Clustering Results}
By simultaneously exploiting the good property of GCRL and taking advantage of the clustering labels, SCAGC ought to learn a discriminative node representation and desirable clustering label at the same time. To illustrate how SCAGC achieves the goal, as shown in Figure~\ref{V-1}, we implement t-SNE~\cite{vandermaaten08a} on the learned $\textbf{M}$ at four different training iterations on ACM and DBLP datasets, where different colors indicated different clustering labels predicted by SCAGC. As observed, the cluster assignments become more reasonable, and different clusters scatter and gather more distinctly. These results indicate that the learned node representation become more compact and discriminative the increasing of the number of iteration.

\subsubsection{Convergence Analysis} Taking ACM dataset as an example, we investigate the convergence of SCAGC. We record the objective values and clustering results of SCAGC with iteration and plot them in Figure~\ref{C-A}. As shown in Figure~\ref{C-A}, the objective values (see the blue line) decrease a lot in the first 100 iterations, then continuously decrease until convergence. Moreover, the ACC of SCAGC continuously increases to a maximum in the first 200 iterations, and generally maintain stable to slight variation. The curves in terms of NMI metric has a similar trend. These observations clearly indicate that SCAGC usually converges quickly.

\section{Conclusion and Future Work}
To conclude, we propose a novel self-supervised contrastive attributed clustering (SCAGC) approach, which can
directly predict the clustering labels of unlabeled attributed graph and handle out-of-sample nodes. We also propose a new self-supervised contrastive loss based on imprecise clustering label to improve the quality of node representation. We believe that the proposed SCAGC will help facilitate the exploration of attributed graph where labels are time and labor consuming to acquire. In the future, we will study how to better explore reliable information embedded in imprecise clustering labels and use it to improve the contrastive loss.

\medskip
{\small
\bibliographystyle{ieee}
\bibliography{egbib}
}

\end{document}